\documentclass[10pt,letterpaper]{article}
\usepackage[top=0.85in,left=1.0in,footskip=0.75in]{geometry}
\usepackage{graphicx}
\usepackage{placeins}

\usepackage{amsmath,amssymb}

\usepackage{changepage}

\usepackage[utf8x]{inputenc}

\usepackage{textcomp,marvosym}

\usepackage{cite}

\usepackage{nameref,hyperref}

\usepackage[right]{lineno}

\usepackage{microtype}
\DisableLigatures[f]{encoding = *, family = * }

\usepackage[table]{xcolor}

\usepackage{array}

\newcolumntype{+}{!{\vrule width 2pt}}

\newlength\savedwidth

\raggedright
\usepackage[aboveskip=1pt,labelfont=bf,labelsep=period,justification=raggedright,singlelinecheck=off]{caption}

\makeatletter
\renewcommand{\@biblabel}[1]{\quad#1.}
\makeatother

\date{}

\usepackage{url}
\usepackage{pbox}
\usepackage{siunitx,booktabs}
\usepackage{color,soul}      

\newcommand{\favg}{$\overline{F_{1}}$}
\newcommand{\alfa}{$\mathit{Alpha}$}
\begin{document}


\begin{flushleft}
{\Large
\textbf\newline{How to evaluate sentiment classifiers for Twitter time-ordered data?}
}
\newline
\\
Igor Mozeti\v{c}\textsuperscript{1*},
Luis Torgo\textsuperscript{2,3},
Vitor Cerqueira\textsuperscript{2},
Jasmina Smailovi\'{c}\textsuperscript{1}
\\
\bigskip
\textsuperscript{1} Department of Knowledge Technologies, Jo\v{z}ef Stefan Institute, Ljubljana, Slovenia
\\
\textsuperscript{2} INESC TEC, Porto, Portugal
\\
\textsuperscript{3} Faculty of Sciences, University of Porto, Porto, Portugal
\\
\bigskip

* igor.mozetic@ijs.si

\end{flushleft}


\section*{Abstract}
Social media are becoming an increasingly important source of information about the
public mood regarding issues such as elections, Brexit, stock market, etc.
In this paper we focus on sentiment classification of Twitter data.
Construction of sentiment classifiers is a standard text mining task, 
but here we address the question of how to properly evaluate them 
as there is no settled way to do so.
Sentiment classes are ordered and unbalanced, and
Twitter produces a stream of time-ordered data. 
The problem we address concerns the procedures used to obtain reliable 
estimates of performance measures, and whether 
the temporal ordering of the training and test data matters.
We collected a large set of 1.5 million tweets in
13 European languages. We created 138 sentiment models and out-of-sample
datasets, which are used as a gold standard for evaluations.
The corresponding 138 in-sample datasets are used to empirically
compare six different estimation procedures:
three variants of cross-validation, and three variants of sequential 
validation (where test set always follows the training set).
We find no significant difference between the best cross-validation
and sequential validation. However, we observe that all cross-validation
variants tend to overestimate the performance, while the sequential
methods tend to underestimate it. Standard cross-validation with random
selection of examples is significantly worse than the blocked
cross-validation, and should not be used to evaluate classifiers in
time-ordered data scenarios.


\section{Introduction}
\label{sec:intro}

Online social media are becoming increasingly important in our society.
Platforms such as Twitter and Facebook influence the daily lives of people 
around the world. 
Their users create and exchange a wide variety of contents on social media, 
which presents a valuable 
source of information about public sentiment regarding social, 
economic or political issues. In this context, it is important to develop automatic methods 
to retrieve and analyze information from social media.

In the paper we address the task of sentiment analysis of Twitter data. 
The task encompasses identification and categorization of opinions 
(e.g., negative, neutral, or positive) written in quasi-natural language 
used in Twitter posts. 
We focus on estimation procedures of the predictive performance of machine learning 
models used to address this task.
Performance estimation procedures are key to understand the generalization ability of 
the models 
since they present approximations of how these models will behave on unseen data.
In the particular case of sentiment analysis of Twitter data, high volumes of content 
are continuously being generated and there is no immediate feedback about the 
true class of instances. In this context, it is fundamental to adopt appropriate estimation procedures 
in order to get reliable estimates about the performance of the models.

The complexity of Twitter data raises some challenges on how to perform such estimations, as, 
to the best of our knowledge, there is currently no settled approach to this. 
Sentiment classes are typically ordered and unbalanced, and the data itself is time-ordered.
Taking these properties into account is important for the selection of 
appropriate estimation procedures.

The Twitter data shares some characteristics of time series and some of static data. 
A time series is an array of observations at regular or equidistant time points, 
and the observations are in general dependent on previous observations~\cite{anderson1995}. 
On the other hand, Twitter data is time-ordered, but the observations are short texts 
posted by Twitter users at any time and frequency. It can be assumed that original 
Twitter posts are not directly dependent on previous posts. 
However, there is a potential indirect dependence,
demonstrated in important trends and events, through influential users and communities, 
or individual user's habits. These long-term topic drifts are typically not taken into
account by the sentiment analysis models.

We study different performance estimation procedures for sentiment analysis in Twitter data. 
These estimation procedures are based on (\textbf{i}) cross-validation and 
(\textbf{ii}) sequential approaches typically adopted for time series data. 
On one hand, cross-validations explore all the available data, 
which is important for the robustness of estimates. 
On the other hand, sequential approaches are more realistic in the sense that 
estimates are computed on a subset of data always subsequent to the data used 
for training, which means that they take time-order into account.

Our experimental study is performed on a large collection of nearly 1.5 million Twitter posts, 
which are domain-free and in 13 different languages. A realistic scenario is emulated 
by partitioning the data into 138 datasets by language and time window. 
Each dataset is split into an in-sample (a training plus test set), where estimation 
procedures are applied to approximate the performance of a model, and an out-of-sample 
used to compute the gold standard.
Our goal is to understand the ability of each estimation procedure to approximate 
the true error incurred by a given model on the out-of-sample data.

The paper is structured as follows. 
\nameref{sec:relatedWork} provides an overview of the state-of-the-art in estimation methods. 
In section~\nameref{sec:methods} we describe the experimental setting for an 
empirical comparison of estimation procedures for sentiment classification of 
time-ordered Twitter data. We describe the Twitter sentiment datasets,
a machine learning algorithm we employ, performance measures, and how the gold standard 
and estimation results are produced. 
In section~\nameref{sec:results} we present and discuss the results of comparisons 
of the estimation procedures along several dimensions. 
\nameref{sec-conclusions} provide the limitations of our work and give directions for the future.

\section{Related work}
\label{sec:relatedWork}

In this section we briefly review typical estimation methods used in sentiment 
classification of Twitter data. In general, for time-ordered data, the estimation
methods used are variants of cross-validation, or are derived from the methods
used to analyze time series data. We examine the state-of-the-art of these estimation 
methods, pointing out their advantages and drawbacks.

Several works in the literature on sentiment classification of Twitter data employ 
standard cross-validation procedures to estimate the performance of sentiment classifiers.
For example, Agarwal et al. \cite{agarwal2011sentiment} and Mohammad et al.
\cite{mohammad2013nrc} propose different methods for sentiment analysis of Twitter data 
and estimate their performance using 5-fold and 10-fold cross-validation, respectively.
Bermingham and Smeaton \cite{bermingham2010classifying} produce a comparative study 
of sentiment analysis between blogs and Twitter posts, where models are compared 
using 10-fold cross-validation. 
Saif et al. \cite{saif2013evaluation} asses binary classification performance of 
nine Twitter sentiment datasets by 10-fold cross validation.
Other, similar applications of cross-validation 
are given in \cite{saif2012semantic,wang2011topic}. 

On the other hand,
there are also approaches that use methods typical for time series data.
For example, Bifet and Frank \cite{Bifet2010} use the prequential (predictive sequential) 
method to evaluate a sentiment classifier on a stream of Twitter posts. 
Moniz et al. \cite{Moniz2014} present a method for predicting the popularity of news from
Twitter data and sentiment scores, and estimate its performance using a sequential 
approach in multiple testing periods.

The idea behind the $K$-fold cross-validation is to randomly shuffle the data and split it 
in $K$ equally-sized folds. Each fold is a subset of the data randomly picked for testing. 
Models are trained on the $K-1$ folds and their performance is estimated on the left-out fold.
$K$-fold cross-validation has several practical advantages, such as an efficient use of 
all the data. However, it is also based on an assumption that the data is independent 
and identically distributed \cite{arlot2010survey} which is often not true.
For example, in time-ordered data, such as Twitter posts, the data are to some
extent dependent due to the underlying temporal order of tweets. 
Therefore, using $K$-fold cross-validation means that one uses future information 
to predict past events, which might hinder the generalization ability of models. 

There are several methods in the literature designed to cope with dependence 
between observations. The most common are sequential approaches typically used 
in time series forecasting tasks. Some variants of $K$-fold cross-validation 
which relax the independence assumption were also proposed.
For time-ordered data, an estimation procedure is sequential when testing is always 
performed on the data subsequent to the training set. Typically, the data is split into 
two parts, where the first is used to train the model and the second is held out for testing. 
These approaches are also known in the literature as the out-of-sample methods
\cite{tashman2000,Bergmeir2012}.

Within sequential estimation methods one can adopt different strategies regarding 
train/test splitting, growing or sliding window setting, and eventual update of the models. 
In order to produce reliable estimates and test for robustness, Tashman \cite{tashman2000}
recommends employing these strategies in multiple testing periods. 
One should either create groups of data series according to, for example, 
different business cycles \cite{fildes1989evaluation}, or adopt a randomized 
approach, such as in~\cite{torgo2014infra}. 
A more complete overview of these approaches is given by Tashman \cite{tashman2000}.

In stream mining, where a model is continuously updated, the most commonly used estimation
methods are holdout and prequential \cite{bifet09,ikonomovska2011learning}. 
The prequential strategy uses an incoming observation to first test the model 
and then to train it.

Besides sequential estimation methods, some variants of $K$-fold cross-validation 
were proposed in the literature that are specially designed to cope with dependency 
in the data and enable the application of cross-validation to time-ordered data. 
For example, blocked cross-validation (the name is adopted from Bergmeir \cite{Bergmeir2012}) 
was proposed by Snijders \cite{snijders1988cross}. 
The method derives from a standard $K$-fold cross-validation, but there is no initial 
random shuffling of observations. This renders $K$ blocks of contiguous observations.

The problem of data dependency for cross-validation is addressed by McQuarrie and Tsai
\cite{mcquarrie1998regression}. The modified cross-validation removes observations 
from the training set that are dependent with the test observations. 
The main limitation of this method is its inefficient use of the available data 
since many observations are removed, as pointed out in \cite{bergmeir2015note}.
The method is also known as non-dependent cross-validation \cite{Bergmeir2012}.

The applicability of variants of cross-validation methods in time series data,
and their advantages over traditional sequential validations are corroborated
by Bergmeir et al. \cite{bergmeir2011forecaster,Bergmeir2012,bergmeir2014}. 
The authors conclude that in time series forecasting tasks, the blocked cross-validations
yield better error estimates because of their more efficient use of the available data. 
Cerqueira et al. \cite{Cerqueira2017comparative} compare performance estimation of various 
cross-validation and out-of-sample approaches on real-world and synthetic time series data. 
The results indicate that cross-validation is appropriate for the stationary synthetic 
time series data, while the out-of-sample approaches yield better estimates for real-world data.

Our contribution to the state-of-the-art is a large scale empirical 
comparison of several estimation procedures on Twitter sentiment data. 
We focus on the differences between the
cross-validation and sequential validation methods, to see how important is the
violation of data independence in the case of Twitter posts. We consider 
longer-term time-dependence between the training and test sets, 
and completely ignore finer-scale dependence at the level of individual tweets 
(e.g., retweets and replies). To the best of our knowledge, there is no settled approach 
yet regarding proper validation of models for Twitter time-ordered data. 
This work provides some results which contribute to bridging that gap.

\section{Methods and experiments}
\label{sec:methods}

The goal of this study is to recommend appropriate estimation procedures
for sentiment classification of Twitter time-ordered data.
We assume a static sentiment classification model applied to
a stream of Twitter posts. In a real-case scenario, the model is
trained on historical, labeled tweets, and applied to the current,
incoming tweets. We emulate this scenario by exploring a large
collection of nearly 1.5 million manually labeled tweets in 13 European languages
(see subsection \nameref{sec:data}).
Each language dataset is split into pairs of the in-sample data, on which a model
is trained, and the out-of-sample data, on which the model is validated.
The performance of the model on the out-of-sample data gives an estimate
of its performance on the future, unseen data. Therefore, we first compute 
a set of 138 out-of-sample performance results, to be used as a gold standard
(subsection \nameref{sec:gold}).
In effect, our goal is to find the estimation procedure that best approximates 
this out-of-sample performance. 

Throughout our experiments we use only one training algorithm (subsection \nameref{sec:data}),
and two performance measures (subsection \nameref{sec:measures}). 
During training, the performance of the trained model
can be estimated only on the in-sample data. However, there are different
estimation procedures which yield these approximations. In machine learning,
a standard procedure is cross-validation, while for time-ordered data,
sequential validation is typically used. In this study, we compare three
variants of cross-validation and three variants of sequential validation
(subsection \nameref{sec:eval-proc}).
The goal is to find the in-sample estimation procedure that best approximates
the out-of-sample gold standard. The error an estimation procedure makes
is defined as the difference to the gold standard.

\subsection{Data and models}
\label{sec:data}

We collected a large corpus of nearly 1.5 million Twitter posts written 
in 13 European languages. This is, to the best of our knowledge, by far the largest 
set of sentiment labeled tweets publicly available.
We engaged native speakers to label the tweets
based on the sentiment expressed in them. The sentiment label has three
possible values: negative, neutral or positive. It turned out that
the human annotators perceived the values as ordered.
The quality of annotations varies though,
and is estimated from the self- and inter-annotator agreements.
All the details about the datasets, the annotator agreements, and the ordering
of sentiment values are in our previous study~\cite{mozetic2016multilingual}.
The sentiment distribution and quality of individual language datasets is in
Table~\ref{tab:DatasetsLabelDistribution}.
The tweets in the datasets are ordered by tweet ids, which corresponds to ordering 
by the time of posting.

\begin{table}[h!]
\centering
\caption{Sentiment label distribution of Twitter datasets in 13 languages.
The last column is a qualitative assessment of the annotation quality,
based on the levels of the self- and inter-annotator agreement.}
    \begin{tabular}{ll|rrr|rl}
    \hline
    Language & & Negative & Neutral & Positive & Total & Quality \\ 
    \hline
    Albanian & alb & 7,062  & 15,066 & 23,630 & 45,758 & poor\\
    Bulgarian & bul & 14,374 & 28,961 & 19,932 & 63,267 & fair \\
    English & eng & 23,250 & 38,457 & 25,721 & 87,428 & v.good \\
    German & ger & 19,039 & 52,166 & 26,743 & 97,948 & fair \\
    Hungarian & hun & 9,062  & 17,833 & 30,410 & 57,305 & good \\
    Polish & pol & 59,027 & 48,658 & 84,245 & 191,930 & good \\
    Portuguese & por & 56,008 & 53,026 & 43,009 & 152,043 & fair \\
    Russian & rus & 30,249 & 37,401 & 25,671 & 93,321 & good \\
    Ser/Cro/Bos & scb & 58,796 & 61,265 & 73,766 & 193,827 & fair \\
    Slovak & slk& 15,060 & 13,112 & 30,598 & 58,770 & good \\
    Slovenian & slv & 34,164 & 48,458 & 30,210 & 112,832 & good \\
    Spanish & spa & 27,675 & 88,481 & 117,048 & 233,204 & poor \\
    Swedish & swe & 22,381 & 15,387 & 13,630 & 51,398 & good \\
    \hline
    Total & & 376,147  & 518,271  & 544,613 & 1,439,031 & \\
    \hline
    \end{tabular}
\label{tab:DatasetsLabelDistribution}%
\end{table}

There are many supervised machine learning algorithms suitable for training 
sentiment classification models from labeled tweets.
In this study we use a variant of Support Vector Machine (SVM) \cite{Vapnik1995}.
The basic SVM is a two-class, binary classifier.
In the training phase, SVM constructs a hyperplane in a high-dimensional 
vector space that separates one class from the other.
In the classification phase, the side of the hyperplane determines the class.
A two-class SVM can be extended into a multi-class classifier which
takes the ordering of sentiment values into account, and
implements ordinal classification \cite{gaudette2009evaluation}. 
Such an extension consists of two SVM classifiers: 
one classifier is trained to separate the negative examples from the
neutral-or-positives; the other separates the negative-or-neutrals 
from the positives.
The result is a classifier with two hyperplanes, which partitions the vector space into 
three subspaces: negative, neutral, and positive. During classification,
the distances from both hyperplanes determine the predicted class.
A further refinement is a \textbf{TwoPlaneSVMbin} classifier.
It partitions the space around both hyperplanes into bins,
and computes the distribution of the training examples in individual bins.
During classification, the distances from both hyperplanes determine the appropriate 
bin, but the class is determined as the majority class in the bin.

The vector space is defined by the features extracted from the Twitter posts.
The posts are first pre-processed by standard text processing methods, 
i.e., tokenization, stemming/lemmatization (if available for a specific language),
unigram and bigram construction, and elimination of terms that do not appear
at least 5 times in a dataset. 
The Twitter specific pre-processing is then applied, i.e, replacing URLs, 
Twitter usernames and hashtags with common tokens, adding emoticon features 
for different types of emoticons in tweets, handling of repetitive letters, etc. 
The feature vectors are then constructed by the Delta TF-IDF weighting scheme 
\cite{martineau2009delta}.

In our previous study \cite{mozetic2016multilingual} we compared five
variants of the SVM classifiers and Naive Bayes on the Twitter
sentiment classification task. TwoPlaneSVMbin was always between the top,
but statistically indistinguishable, best performing classifiers.
It turned out that monitoring the quality of the annotation
process has much larger impact on the performance than the type of
the classifier used. In this study we fix the classifier, and use
TwoPlaneSVMbin in all the experiments.

\subsection{Performance measures}
\label{sec:measures}

Sentiment values are ordered, and distribution of tweets between the
three sentiment classes is often unbalanced.
In such cases, \textit{accuracy} is not the most appropriate performance 
measure \cite{Bifet2010,mozetic2016multilingual}.
In this context, we evaluate performance with the following two metrics: 
Krippendorff's \alfa\, \cite{Krippendorff2012},
and \favg\, \cite{kiritchenko2014sentiment}.

\alfa\, was developed to measure the agreement between human annotators,
but can also be used to measure the agreement between classification models
and a gold standard. It generalizes several specialized agreement measures,
takes ordering of classes into account, and accounts for the agreement by chance.
\alfa\, is defined as follows:
\begin{equation}
 \mathit{Alpha} = 1 - \frac{D_{o}}{D_{e}} \,
\end{equation}
where $D_{o}$ is the observed disagreement between models, and
$D_{e}$ is a disagreement, expected by chance.
When models agree perfectly, \alfa\;$=1$, and when the level of 
agreement equals the agreement by chance, \alfa\;$=0$. 
Note that \alfa\, can also be negative.
The two disagreement measures are defined as:
\begin{equation}
D_{o} = \frac{1}{N} \sum_{c,c'} N(c,c') \cdot \delta^2(c,c') \,,
\end{equation}
\begin{equation}
D_{e} = \frac{1}{N(N-1)} \sum_{c,c'} N(c) \cdot N(c') \cdot \delta^2(c,c') \,.
\end{equation}
The arguments, $N, N(c,c'), N(c)$, and $N(c')$,
refer to the frequencies in a coincidence matrix, defined below.
$c$ (and $c'$) is a discrete sentiment variable with three possible values:
\textit{negative} ($-1$), \textit{neutral} (0), or \textit{positive} ($+1$).
$\delta(c,c')$ is a difference function between the values of $c$ and $c'$,
for ordered variables defined as:
\begin{equation}
\delta(c,c') = |c - c'| \;\;\;\; c,c'\in \{-1,0,+1\} \;.
\end{equation}
Note that disagreements $D_{o}$ and $D_{e}$ between the extreme 
classes (\textit{negative} and \textit{positive})
are four times larger than between the neighbouring classes.

A coincidence matrix tabulates all pairable values of $c$ from two models.
In our case, we have a $3$-by-$3$ coincidence matrix, and compare
a model to the gold standard. The coincidence
matrix is then the sum of the confusion matrix and its transpose.
Each labeled tweet is entered twice,
once as a $(c,c')$ pair, and once as a $(c',c)$ pair.
$N(c,c')$ is the number of tweets labeled by the values $c$ and $c'$
by different models, $N(c)$ and $N(c')$ are the totals
for each value, and $N$ is the grand total.

\favg\, is an instance of the $F$ score, a well-known performance measure 
in information retrieval \cite{VanRijsbergen1979} and machine learning.
We use an instance specifically designed to evaluate the 3-class 
sentiment models \cite{kiritchenko2014sentiment}.
\favg\, is defined as follows:
\begin{equation}
\overline{F_1} = \frac{F_1(-1) + F_1(+1)}{2} \,.
\end{equation}
\favg\, implicitly takes into account the ordering of sentiment values,
by considering only the extreme labels, \textit{negative} $(-1)$ and \textit{positive} $(+1)$.
The middle, \textit{neutral}, is taken into account only indirectly.
$F_{1}(c)$ is the harmonic mean of precision and recall for class $c$,
$c\in \{-1,+1\}$.
\favg\;$=1$ implies that all negative and positive tweets were correctly classified,
and as a consequence, all neutrals as well. \favg\;$=0$ indicates that all negative
and positive tweets were incorrectly classified. \favg\, does not account for
correct classification by chance.

\subsection{Gold standard}
\label{sec:gold}

\begin{figure*}[h!]
\begin{center}
\includegraphics[width=15cm]{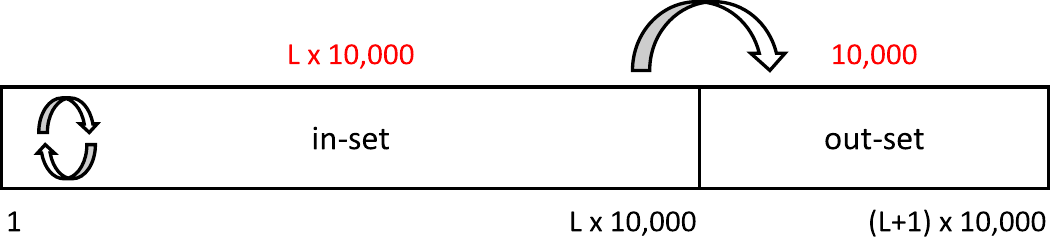}
\caption{Creation of the estimation and gold standard data.
Each labeled language dataset (Table~\ref{tab:DatasetsLabelDistribution})
is partitioned into $L$ in-sets and corresponding out-sets.
The in-sets always start at the first tweet and are progressively longer
in multiples of 10,000 tweets.
The corresponding out-set is the subsequent 10,000 consecutive tweets, 
or the remainder at the end of the language dataset.}
\label{fig:InsetOutsetPartitioning}
\end{center}
\end{figure*}

We create the gold standard results by splitting the data into the in-sample datasets
(abbreviated as in-set), and out-of-sample datasets (abbreviated as out-set).
The terminology of the in- and out-set is adopted from Bergmeir et al.~\cite{Bergmeir2012}.
Tweets are ordered by the time of posting.
To emulate a realistic scenario, an out-set always follows the in-set.
From each language dataset (Table~\ref{tab:DatasetsLabelDistribution}) 
we create $L$ in-sets of varying length in multiples of 10,000 consecutive tweets, 
where $L = \lfloor N/10000 \rfloor$.
The out-set is the subsequent 10,000 consecutive tweets, or the remainder at
the end of each language dataset. This is illustrated in Figure~\ref{fig:InsetOutsetPartitioning}.

The partitioning of the language datasets results in 138 in-sets and
corresponding out-sets. For each in-set, we train a TwoPlaneSVMbin sentiment
classification model, and measure its performance, in terms of \alfa\, and \favg,
on the corresponding out-set.
The results are in Tables~\ref{tab:realWorldScenarioAlpha} and \ref{tab:realWorldScenarioF1}.
Note that the performance measured by \alfa\, is considerably lower in
comparison to \favg, since the baseline for \alfa\, is classification by chance.

\begin{table}[h!]
\centering
\caption{Gold standard performance results as measured by \alfa.
The baseline, \alfa$=0$, indicates classification by chance.}
  \resizebox{\textwidth}{!}{
    \begin{tabular}{rrrrrrrrrrrrr}
    \hline
    alb & bul & eng & ger & hun & pol & por & rus & scb & slk & slv & spa & swe \\
    \hline
    0.210 & 0.321 & 0.414 & 0.391 & 0.419 & 0.409 & 0.338 & 0.369 & 0.275 & 0.367 & 0.327 & 0.171 & 0.470 \\
    0.102 & 0.324 & 0.433 & 0.420 & 0.453 & 0.432 & 0.336 & 0.420 & 0.393 & 0.411 & 0.380 & 0.222 & 0.463 \\
    0.084 & 0.339 & 0.449 & 0.423 & 0.482 & 0.479 & 0.360 & 0.441 & 0.408 & 0.425 & 0.414 & 0.255 & 0.458 \\
    0.106 & 0.363 & 0.474 & 0.416 & 0.460 & 0.499 & 0.428 & 0.435 & 0.457 & 0.438 & 0.439 & 0.269 & 0.473 \\
          & 0.375 & 0.513 & 0.387 & 0.475 & 0.486 & 0.183 & 0.478 & 0.421 & 0.454 & 0.453 & 0.211 & 0.480 \\
          & 0.397 & 0.513 & 0.403 &       & 0.487 & 0.176 & 0.452 & 0.327 &       & 0.478 & 0.227 &  \\
          &       & 0.541 & 0.406 &       & 0.483 & 0.224 & 0.492 & 0.293 &       & 0.455 & 0.226 &  \\
          &       & 0.526 & 0.354 &       & 0.512 & 0.333 & 0.474 & 0.341 &       & 0.418 & 0.227 &  \\
          &       &       & 0.351 &       & 0.467 & 0.388 & 0.489 & 0.358 &       & 0.425 & 0.151 &  \\
          &       &       &       &       & 0.513 & 0.409 &       & 0.384 &       & 0.418 & 0.193 &  \\
          &       &       &       &       & 0.491 & 0.425 &       & 0.382 &       & 0.320 & 0.196 &  \\
          &       &       &       &       & 0.526 & 0.434 &       & 0.485 &       &       & 0.220 &  \\
          &       &       &       &       & 0.549 & 0.439 &       & 0.528 &       &       & 0.233 &  \\
          &       &       &       &       & 0.535 & 0.453 &       & 0.551 &       &       & 0.207 &  \\
          &       &       &       &       & 0.541 & 0.472 &       & 0.512 &       &       & 0.202 &  \\
          &       &       &       &       & 0.500 &       &       & 0.533 &       &       & 0.179 &  \\
          &       &       &       &       & 0.544 &       &       & 0.418 &       &       & 0.159 &  \\
          &       &       &       &       & 0.532 &       &       & 0.514 &       &       & 0.207 &  \\
          &       &       &       &       & 0.528 &       &       & 0.479 &       &       & 0.216 &  \\
          &       &       &       &       &       &       &       &       &       &       & 0.251 &  \\
          &       &       &       &       &       &       &       &       &       &       & 0.241 &  \\
          &       &       &       &       &       &       &       &       &       &       & 0.110 &  \\
          &       &       &       &       &       &       &       &       &       &       & 0.142 &  \\
    \hline
    \end{tabular}
    }
\label{tab:realWorldScenarioAlpha}
\end{table}

\begin{table}[h!]
\centering
\caption{Gold standard performance results as measured by \favg.
The baseline, \favg$=0$, indicates that all negative and positive examples
are classified incorrectly.}
    \resizebox{\textwidth}{!}{
    \begin{tabular}{rrrrrrrrrrrrr}
    \hline
    alb & bul & eng & ger & hun & pol & por & rus & scb & slk & slv & spa & swe \\
    \hline
    0.479 & 0.509 & 0.545 & 0.578 & 0.610 & 0.621 & 0.356 & 0.551 & 0.492 & 0.616 & 0.485 & 0.436 & 0.627 \\
    0.396 & 0.501 & 0.567 & 0.595 & 0.624 & 0.632 & 0.358 & 0.560 & 0.569 & 0.657 & 0.533 & 0.452 & 0.620 \\
    0.387 & 0.498 & 0.571 & 0.588 & 0.637 & 0.653 & 0.383 & 0.572 & 0.577 & 0.669 & 0.567 & 0.504 & 0.629 \\
    0.388 & 0.510 & 0.595 & 0.561 & 0.628 & 0.670 & 0.449 & 0.571 & 0.626 & 0.670 & 0.593 & 0.473 & 0.630 \\
          & 0.513 & 0.634 & 0.533 & 0.640 & 0.651 & 0.243 & 0.604 & 0.580 & 0.675 & 0.603 & 0.446 & 0.658 \\
          & 0.535 & 0.640 & 0.537 &       & 0.663 & 0.252 & 0.588 & 0.485 &       & 0.624 & 0.454 &  \\
          &       & 0.654 & 0.529 &       & 0.656 & 0.322 & 0.617 & 0.469 &       & 0.550 & 0.440 &  \\
          &       & 0.647 & 0.409 &       & 0.682 & 0.448 & 0.610 & 0.493 &       & 0.521 & 0.438 &  \\
          &       &       & 0.413 &       & 0.654 & 0.529 & 0.614 & 0.503 &       & 0.524 & 0.429 &  \\
          &       &       &       &       & 0.672 & 0.556 &       & 0.526 &       & 0.507 & 0.424 &  \\
          &       &       &       &       & 0.659 & 0.589 &       & 0.573 &       & 0.415 & 0.412 &  \\
          &       &       &       &       & 0.680 & 0.605 &       & 0.654 &       &       & 0.407 &  \\
          &       &       &       &       & 0.696 & 0.608 &       & 0.686 &       &       & 0.431 &  \\
          &       &       &       &       & 0.679 & 0.624 &       & 0.696 &       &       & 0.398 &  \\
          &       &       &       &       & 0.682 & 0.638 &       & 0.665 &       &       & 0.403 &  \\
          &       &       &       &       & 0.650 &       &       & 0.684 &       &       & 0.402 &  \\
          &       &       &       &       & 0.670 &       &       & 0.644 &       &       & 0.390 &  \\
          &       &       &       &       & 0.663 &       &       & 0.661 &       &       & 0.446 &  \\
          &       &       &       &       & 0.663 &       &       & 0.625 &       &       & 0.479 &  \\
          &       &       &       &       &       &       &       &       &       &       & 0.516 &  \\
          &       &       &       &       &       &       &       &       &       &       & 0.516 &  \\
          &       &       &       &       &       &       &       &       &       &       & 0.423 &  \\
          &       &       &       &       &       &       &       &       &       &       & 0.449 &  \\
    \hline
    \end{tabular}
    }
\label{tab:realWorldScenarioF1}
\end{table}

The 138 in-sets are used to train sentiment classification models and estimate
their performance. The goal of this study is to analyze different estimation
procedures in terms of how well they approximate the out-set gold standard results shown in Tables~\ref{tab:realWorldScenarioAlpha} and \ref{tab:realWorldScenarioF1}.

\FloatBarrier
\subsection{Estimation procedures}
\label{sec:eval-proc}

There are different estimation procedures, some more suitable for static data,
while  others are more appropriate for time-series data.
Time-ordered Twitter data shares some properties of both types of data.
When training an SVM model, the order of tweets is irrelevant and the model
does not capture the dynamics of the data.
When applying the model, however, new tweets might introduce new vocabulary
and topics. As a consequence, the temporal ordering of training and test data
has a potential impact on the performance estimates.

We therefore compare two classes of estimation procedures.
Cross-validation, commonly used in machine learning for model evaluation
on static data, and sequential validation, commonly used for time-series data.
There are many variants and parameters for each class of procedures.
Our datasets are relatively large and an application of each estimation
procedure takes several days to complete.
We have selected three variants of each procedure to provide answers
to some relevant questions.

First, we apply 10-fold cross-validation where the training:test set ratio is always 9:1.
Cross-validation is \textit{stratified} when the fold
partitioning is not completely random, but each fold has roughly the same
class distribution. We also compare standard \textit{random} selection of examples
to the \textit{blocked} form of cross-validation \cite{snijders1988cross,Bergmeir2012},
where each fold is a block of consecutive tweets.
We use the following abbreviations for cross-validations:
\begin{itemize}
\item \textbf{xval(9:1, strat, block)} - 10-fold, stratified, blocked;
\item \textbf{xval(9:1, no-strat, block)} - 10-fold, not stratified, blocked;
\item \textbf{xval(9:1, strat, rand)} - 10-fold, stratified, random selection of examples.
\end{itemize}

\begin{figure*}[h!]
\begin{center}
\includegraphics[width=15cm]{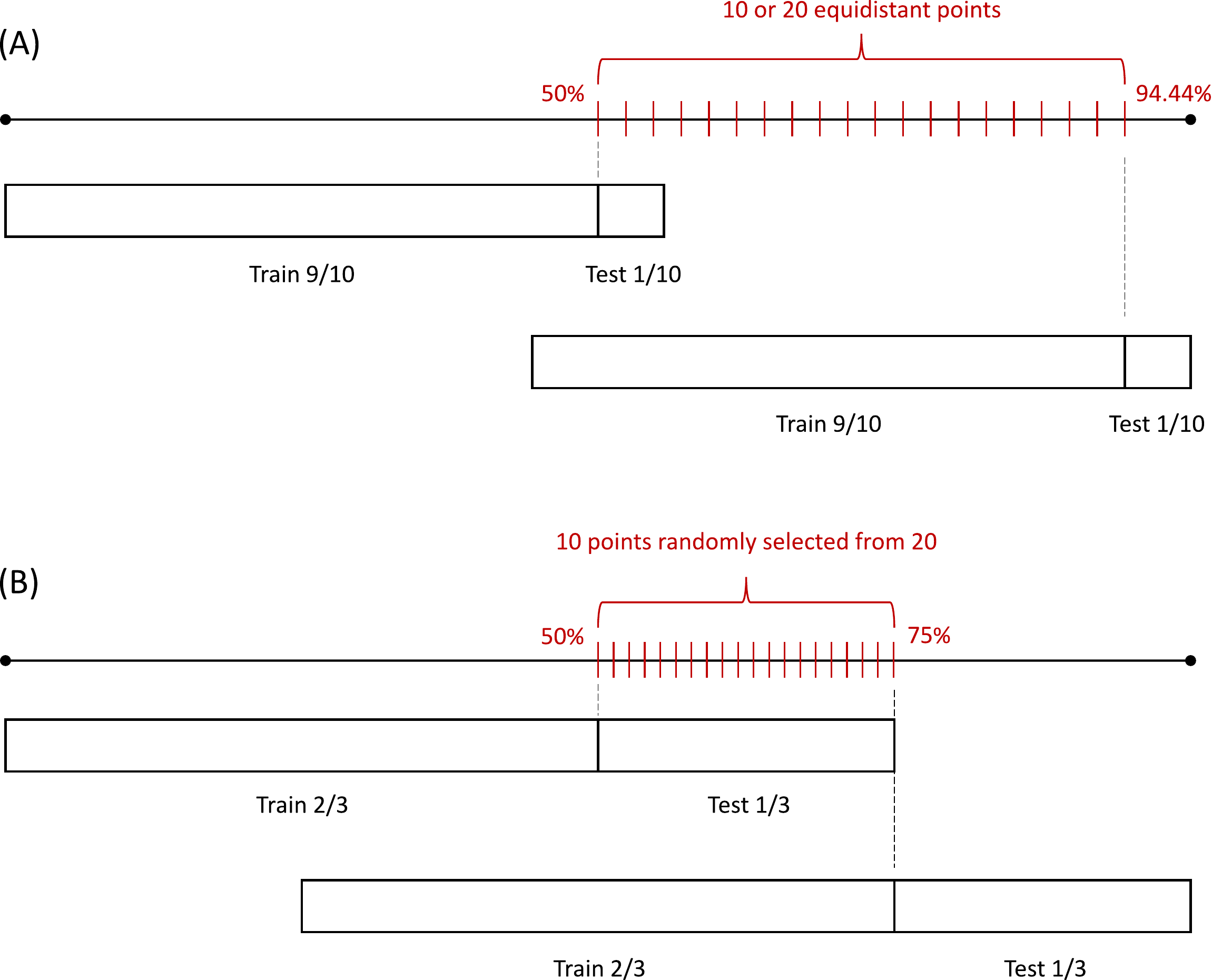}
\caption{Sampling of an in-set for sequential validation.
A sample consists of a training set, immediately followed by a test set.
We consider two scenarios: (A) The ratio of the training and test set is 9:1,
and the sample is shifted along 10 or 20 equidistant points.
(B) The training:test set ratio is 2:1 and the sample is positioned
at 10 randomly selected points out of 20 equidistant points.}
\label{fig:sequentialBlockEvaluationAB}
\end{center}
\end{figure*}

In sequential validation, a sample consists of the training set immediately followed
by the test set.
We vary the ratio of the training and test set sizes, and the number 
and distribution of samples taken from the in-set. The number of samples
is 10 or 20, and they are distributed equidistantly or semi-equidistantly.
In all variants, samples cover the whole in-set, but they are overlapping.
See Figure~\ref{fig:sequentialBlockEvaluationAB} for illustration.
We use the following abbreviations for sequential validations:
\begin{itemize}
\item \textbf{seq(9:1, 20, equi)} - 9:1 training:test ratio, 20 equidistant samples,
\item \textbf{seq(9:1, 10, equi)} - 9:1 training:test ratio, 10 equidistant samples,
\item \textbf{seq(2:1, 10, semi-equi)} - 2:1 training:test ratio, 10 samples randomly 
selected out of 20 equidistant points.
\end{itemize}

\FloatBarrier

\section{Results and discussion}
\label{sec:results}

We compare six estimation procedures in terms of different types of
errors they incur. The error is defined as the difference to the gold standard.
First, the magnitude and sign of the errors show whether a method tends 
to underestimate or overestimate the performance, and by how much 
(subsection \nameref{sec:median-errors}).
Second, relative errors give fractions of small, moderate, and large
errors that each procedure incurs (subsection \nameref{sec:rel-errors}).
Third, we rank the estimation procedures in terms of increasing absolute
errors, and estimate the significance of the overall ranking by the
Friedman-Nemenyi test (subsection \nameref{sec:friedman}).
Finally, selected pairs of estimation procedures are compared by 
the Wilcoxon signed-rank test (subsection \nameref{sec:wilcoxon}).


\subsection{Median errors} 
\label{sec:median-errors}

An estimation procedure estimates the performance\, (abbreviated $Est$) of a model in 
terms of \alfa\, and \favg. The error it incurs is defined as the difference 
to the gold standard performance (abbreviated $Gold$):
$Err = Est - Gold$.
The validation results show high variability of the errors, with skewed distribution 
and many outliers. Therefore, we summarize the errors in terms of their medians and quartiles, 
instead of the averages and variances. 

The median errors of the six estimation procedures are in 
Tables~\ref{tab:MedianErrorAlpha} and~\ref{tab:MedianErrorF1},
measured by \alfa\, and \favg, respectively.

\begin{table}[h!]
\centering
\caption{Median errors, measured by \alfa, for individual language datasets and 
six estimation procedures.}
 \resizebox{\textwidth}{!}{
    \begin{tabular}{l *{6}{S[table-format=-1.3]}}
    \hline
    Lang &
    \pbox{1.9cm}{xval(9:1,\\strat, block)} &
    \pbox{2.5cm}{xval(9:1,\\no-strat, block)}  &
    \pbox{1.8cm}{xval(9:1,\\strat, rand)} &
    \pbox{1.8cm}{seq(9:1,\\20, equi)} &
    \pbox{1.8cm}{seq(9:1,\\10, equi)} &
    \pbox{2.5cm}{seq(2:1,\\10, semi-equi)} \\
\hline
alb&0.052&0.036&0.206&0.001&0.001&0.001\tabularnewline
bul&0.009&0.013&0.046&-0.019&-0.025&-0.043\tabularnewline
eng&-0.016&-0.017&-0.010&-0.040&-0.042&-0.039\tabularnewline
ger&0.037&0.049&0.059&0.009&0.010&0.001\tabularnewline
hun&0.009&0.013&0.025&-0.011&-0.007&-0.007\tabularnewline
pol&0.011&0.016&0.054&-0.020&-0.017&-0.031\tabularnewline
por&-0.048&-0.048&-0.015&-0.040&-0.045&-0.085\tabularnewline
rus&0.008&0.008&0.029&-0.027&-0.029&-0.045\tabularnewline
scb&-0.046&-0.051&0.026&-0.047&-0.043&-0.069\tabularnewline
slk&0.018&0.015&0.055&-0.025&-0.023&-0.039\tabularnewline
slv&0.003&-0.004&0.040&-0.029&-0.026&-0.031\tabularnewline
spa&-0.008&0.031&0.070&0.012&0.011&-0.011\tabularnewline
swe&0.055&0.057&0.106&0.011&0.006&-0.028\tabularnewline
\hline
Median&0.009&0.013&0.046&-0.020&-0.023&-0.031\tabularnewline
\hline
\end{tabular}
}
\label{tab:MedianErrorAlpha}
\end{table}

\begin{table}[h!]
\centering
\caption{Median errors, measured by \favg, for individual language datasets and 
six estimation procedures.}
\resizebox{\textwidth}{!}{%
    \begin{tabular}{l *{6}{S[table-format=-1.3]}}
\hline
    Lang &
    \pbox{1.9cm}{xval(9:1,\\strat, block)} &
    \pbox{2.5cm}{xval(9:1,\\no-strat, block)}  &
    \pbox{1.8cm}{xval(9:1,\\strat, rand)} &
    \pbox{1.8cm}{seq(9:1,\\20, equi)} &
    \pbox{1.8cm}{seq(9:1,\\10, equi)} &
    \pbox{2.5cm}{seq(2:1,\\10, semi-equi)} \\
\hline
alb&0.026&0.016&0.137&-0.014&-0.007&-0.009\tabularnewline
bul&0.020&0.024&0.047&0.003&-0.002&-0.019\tabularnewline
eng&-0.019&-0.020&-0.015&-0.027&-0.027&-0.028\tabularnewline
ger&0.056&0.058&0.072&0.025&0.028&0.014\tabularnewline
hun&0.022&0.022&0.030&-0.006&-0.009&-0.005\tabularnewline
pol&0.013&0.020&0.044&-0.001&0&-0.007\tabularnewline
por&-0.050&-0.045&-0.040&-0.049&-0.056&-0.092\tabularnewline
rus&0.008&0.010&0.025&-0.019&-0.018&-0.021\tabularnewline
scb&-0.034&-0.037&0&-0.030&-0.032&-0.050\tabularnewline
slk&0.005&0.008&0.025&-0.013&-0.015&-0.013\tabularnewline
slv&0.003&0&0.029&-0.022&-0.026&-0.032\tabularnewline
spa&-0.001&0.024&0.060&0.007&0.010&0.012\tabularnewline
swe&0.030&0.037&0.071&0.008&0.006&-0.011\tabularnewline
\hline
Median&0.008&0.016&0.030&-0.013&-0.009&-0.013\tabularnewline
\hline
\end{tabular}
}
\label{tab:MedianErrorF1}
\end{table}

\begin{figure*}[h!]
\begin{center}
\includegraphics[width=\textwidth]{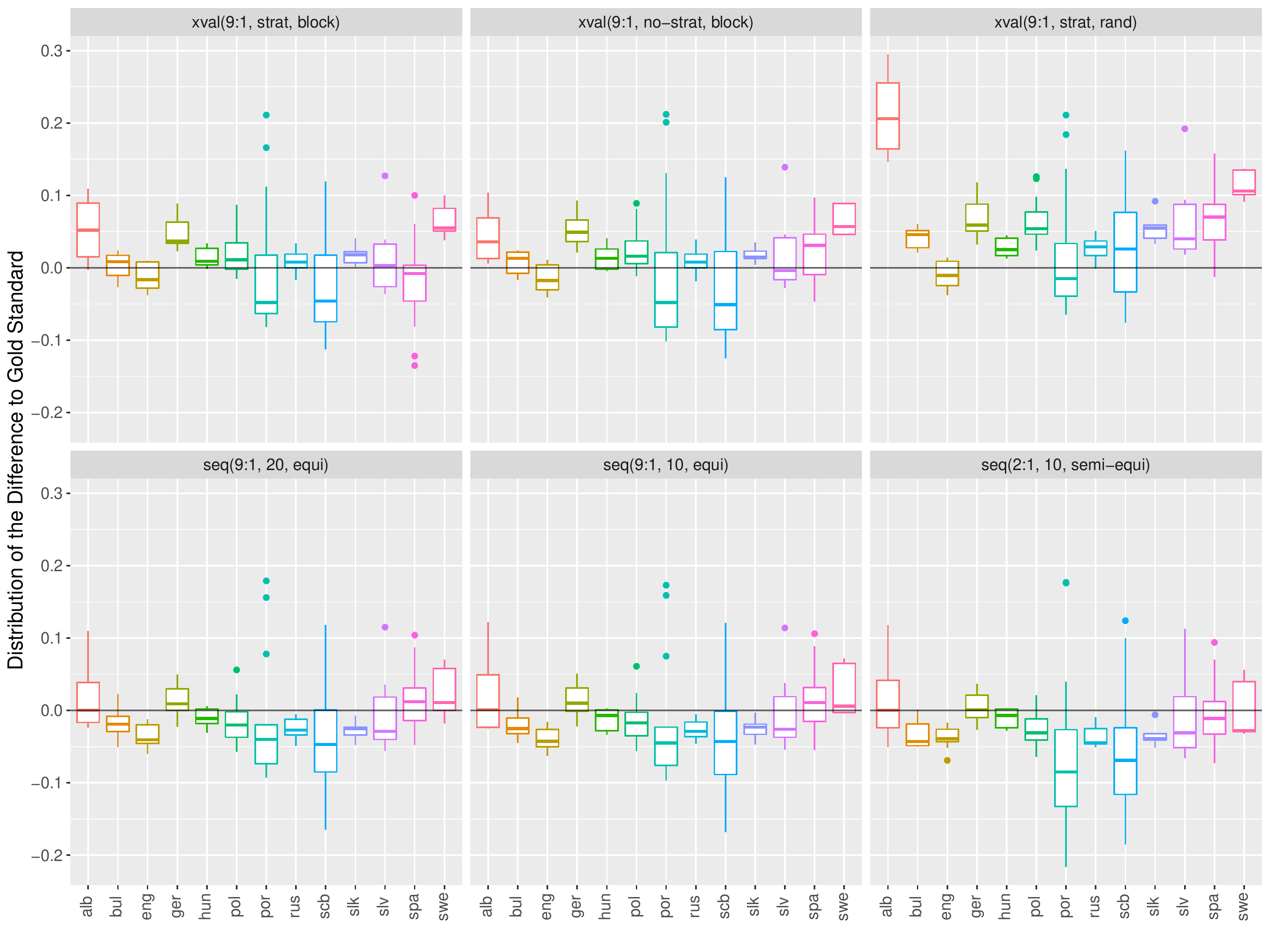}
\caption{Box plots of errors of six estimation procedures for 13 language datasets.
Errors are measured in terms of \alfa.}
\label{fig:Alpha_boxPlotDifs2GS}
\end{center}
\end{figure*}

Figure \ref{fig:Alpha_boxPlotDifs2GS} depicts the errors with box plots.
The band inside the box denotes the median, the box spans the second and third quartile,
and the whiskers denote 1.5 interquartile range. The dots correspond to the outliers.
Figure \ref{fig:Alpha_boxPlotDifs2GS} shows high variability of errors for individual
datasets. This is most pronounced for the Serbian/Croatian/Bosnian (scb) and 
Portuguese (por) datasets where variation in annotation quality (scb) 
and a radical topic shift (por) were observed.
Higher variability is also observed for the Spanish (spa) and Albanian (alb) datasets, 
which have poor sentiment
annotation quality (see \cite{mozetic2016multilingual} for details).

\begin{figure*}[h!]
\begin{center}
\includegraphics[width=\textwidth]{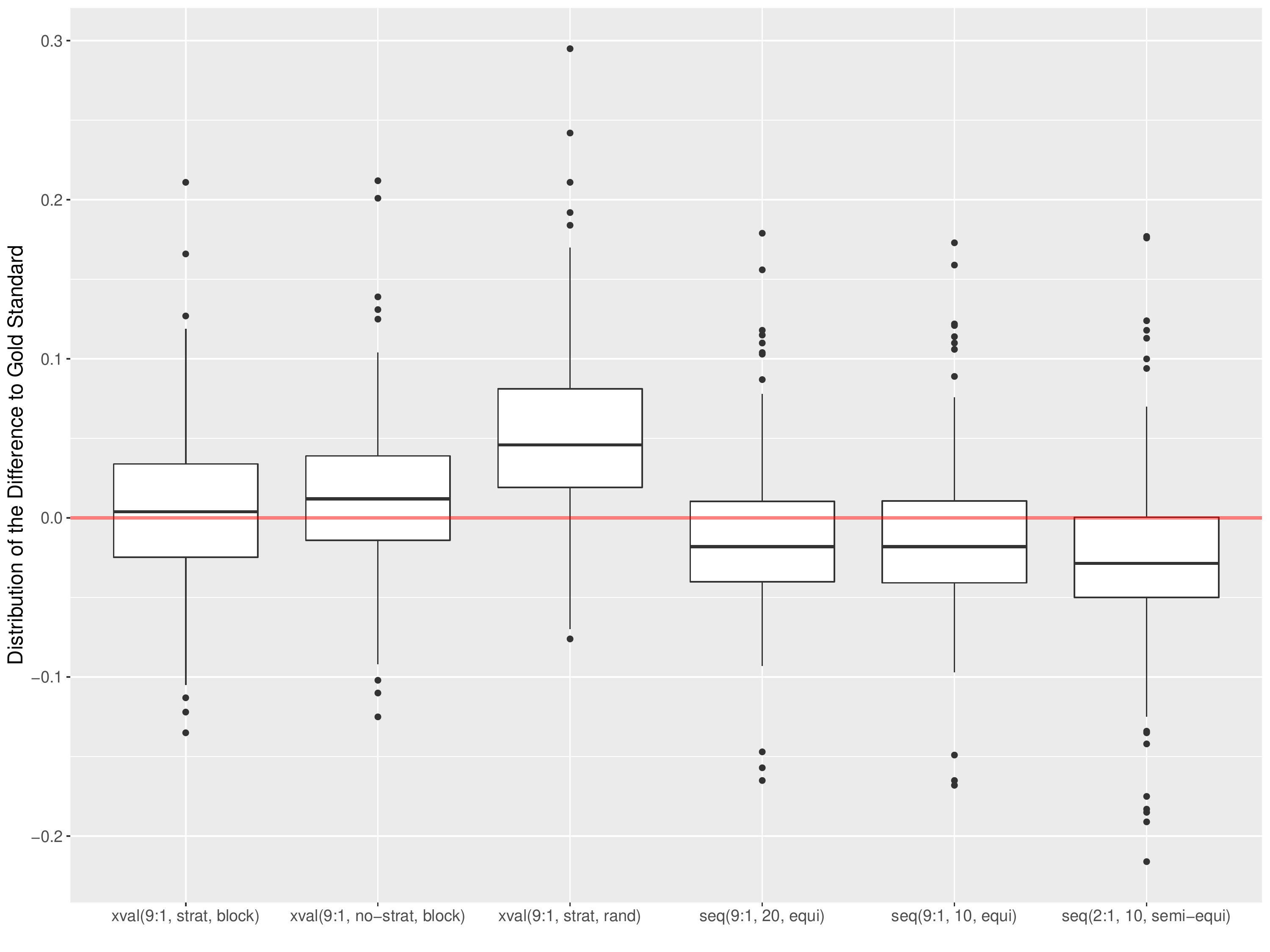}
\caption{Box plots of errors of six estimation procedures aggregated over all 
language datasets. Errors are measured in terms of \alfa.}
\label{fig:Alpha_BoxPlotStrats}
\end{center}
\end{figure*}

\begin{figure*}[h!]
\begin{center}
\includegraphics[width=\textwidth]{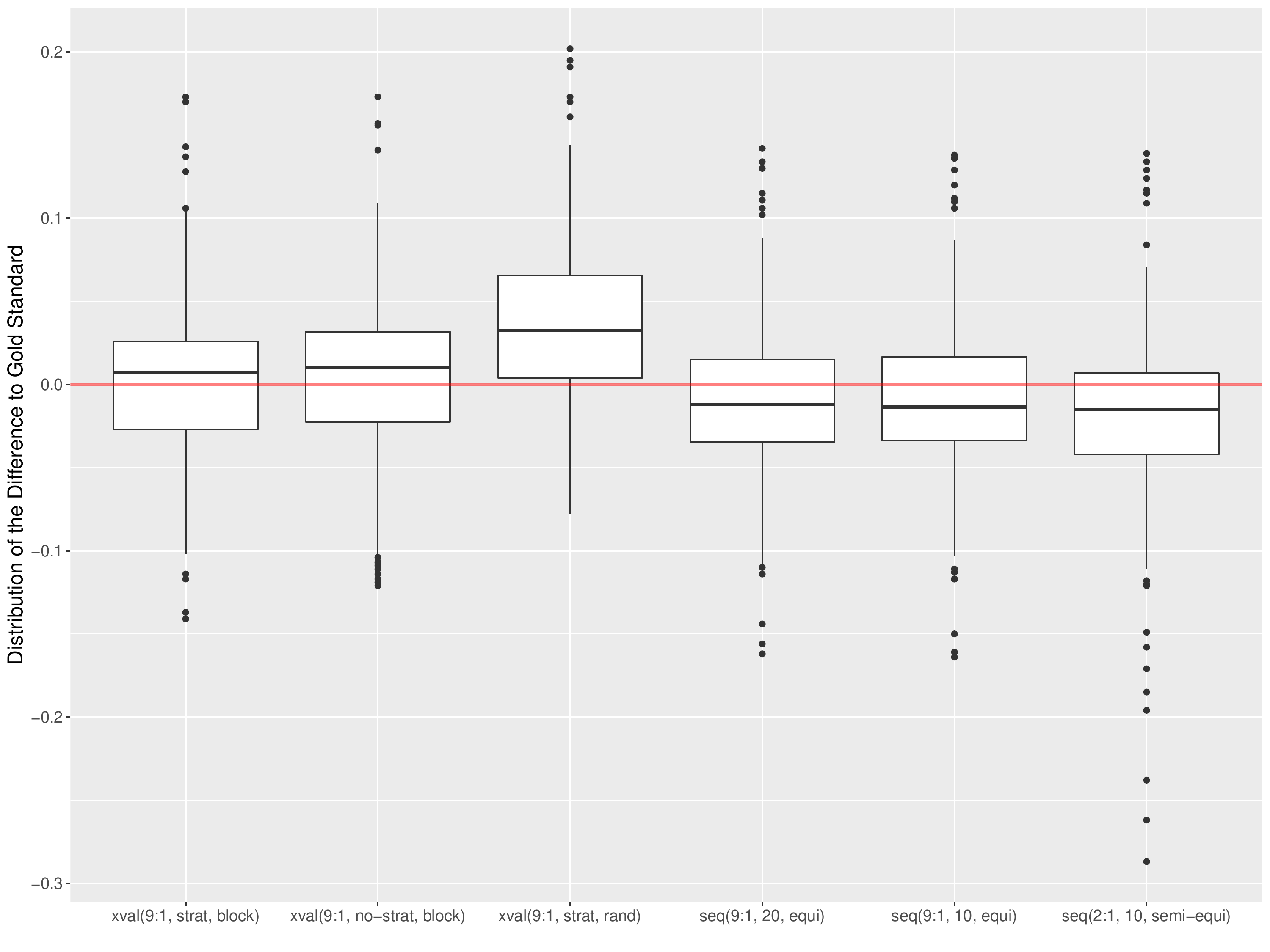}
\caption{Box plots of errors of six estimation procedures aggregated over all 
language datasets. Errors are measured in terms of \favg.}
\label{fig:F_BoxPlotStrats}
\end{center}
\end{figure*}

The differences between the estimation procedures are easier to detect
when we aggregate the errors over all language datasets. The results are in
Figures \ref{fig:Alpha_BoxPlotStrats} and \ref{fig:F_BoxPlotStrats},
for \alfa\, and \favg, respectively.
In both cases we observe that the cross-validation procedures (xval) consistently
overestimate the performance, while the sequential validations (seq) underestimate it.
The largest overestimation errors are incurred by the random cross-validation,
and the largest underestimations by the sequential validation with the training:test
set ratio 2:1. We also observe high variability of errors, with many outliers.
The conclusions are consistent for both measures, \alfa\, and \favg.

\FloatBarrier

\subsection{Relative errors} 
\label{sec:rel-errors}

Another useful analysis of estimation errors is provided by a comparison of
relative errors. The relative error is the absolute error an estimation procedure
incurs divided by the gold standard result:
$RelErr = |Est - Gold| / Gold$.
We chose two, rather arbitrary, thresholds of 5\% and 30\%, and classify the
relative errors as small ($RelErr < 5\%$), moderate ($5\% \le RelErr \le 30\%$), 
and large ($RelErr > 30\%$).

\begin{figure*}[h!]
\begin{center}
\includegraphics[width=\textwidth]{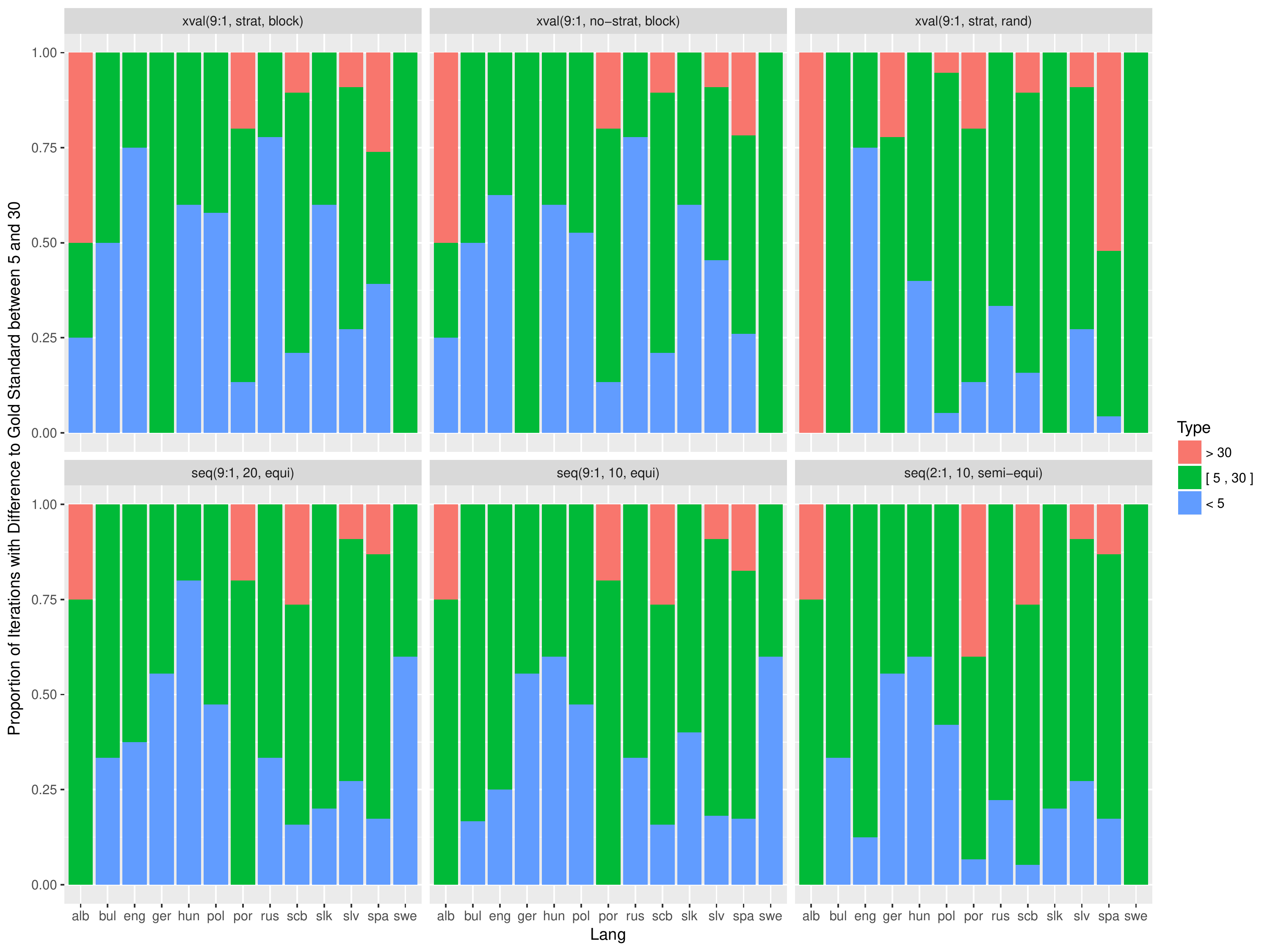}
\caption{Proportion of relative errors, measured by \alfa, per estimation procedure
and individual language dataset. Small errors ($<5\%$) are in blue, 
moderate ($[5,30]\%$) in green, and large errors ($>30\%$) in red.}
\label{fig:Alpha_FailuresBetween5and30}
\end{center}
\end{figure*}

Figure \ref{fig:Alpha_FailuresBetween5and30} shows the proportion of the three
types of errors, measured by \alfa, for individual language datasets. 
Again, we observe a higher proportion
of large errors for languages with poor annotations (alb, spa), 
annotations of different quality (scb), and different topics (por).

\begin{figure*}[h!]
\begin{center}
\includegraphics[width=\textwidth]{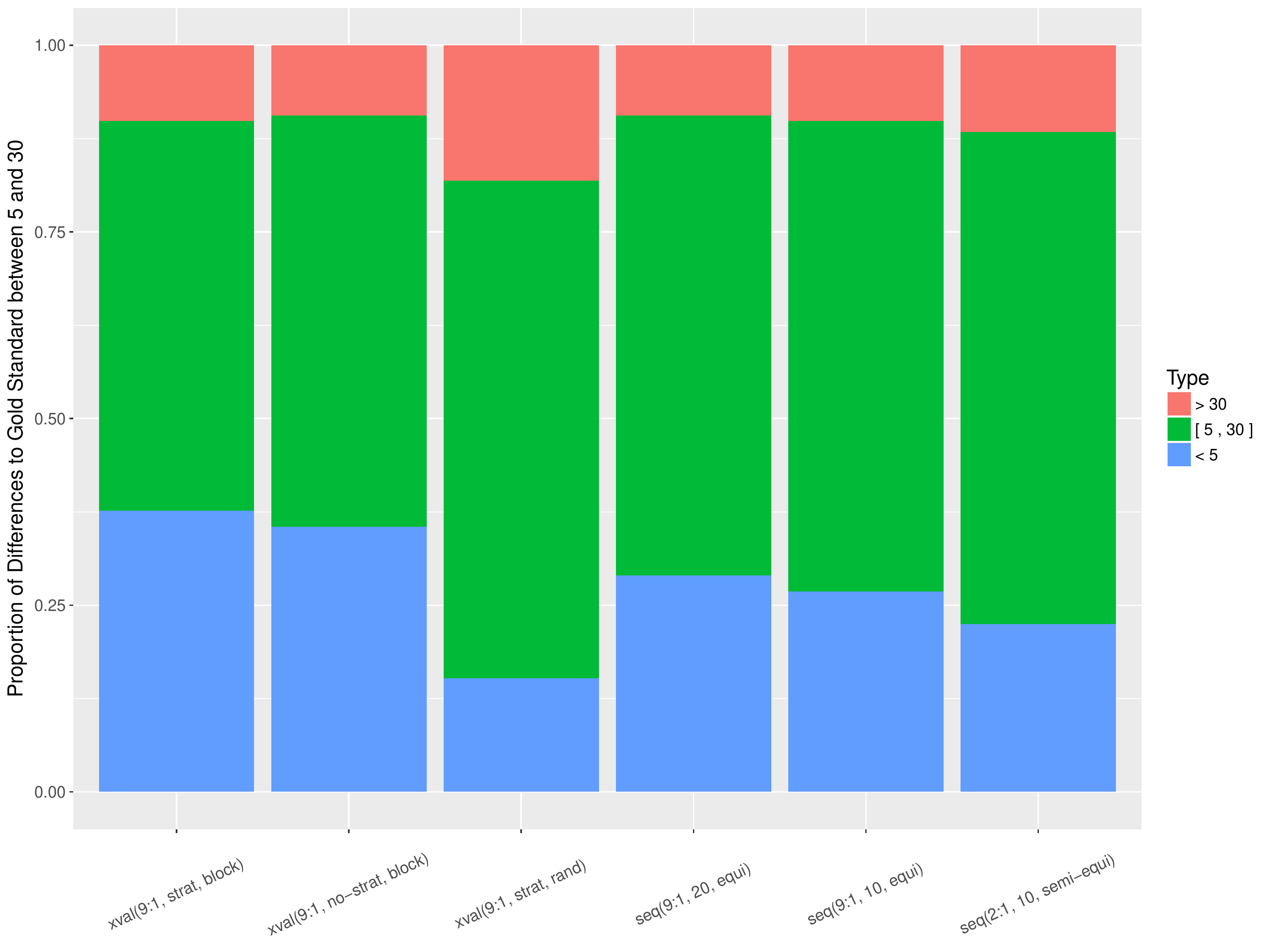}
\caption{Proportion of relative errors, measured by \alfa, per estimation procedure
and aggregated over all 138 datasets. Small errors ($<5\%$) are in blue, 
moderate ($[5,30]\%$) in green, and large errors ($>30\%$) in red.}
\label{fig:Alpha_TotalFailuresBetween5and30}
\end{center}
\end{figure*}

\begin{figure*}[h!]
\begin{center}
\includegraphics[width=\textwidth]{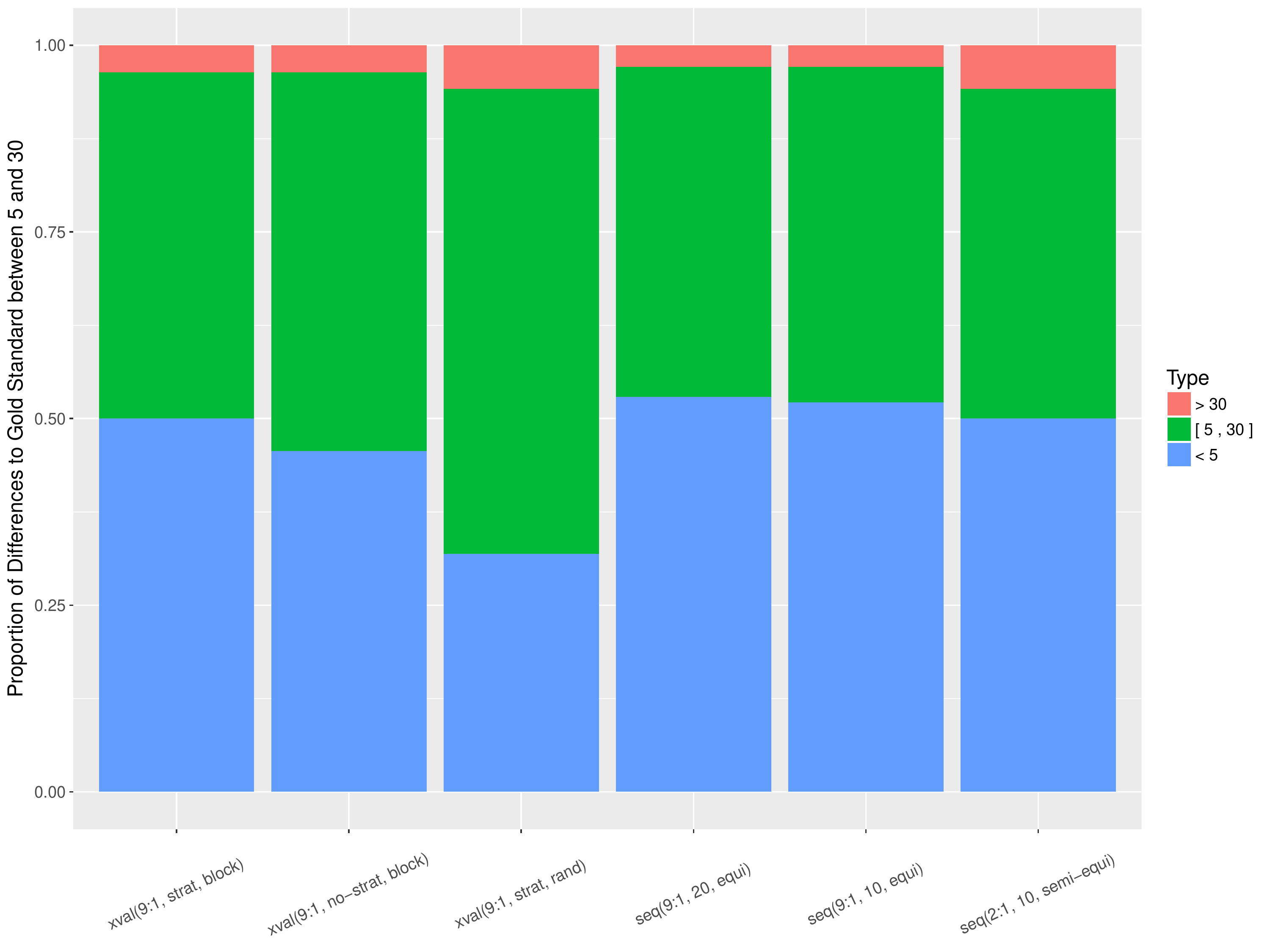}
\caption{Proportion of relative errors, measured by \favg, per estimation procedure
and aggregated over all 138 datasets. Small errors ($<5\%$) are in blue, 
moderate ($[5,30]\%$) in green, and large errors ($>30\%$) in red.}
\label{fig:F_TotalFailuresBetween5and30}
\end{center}
\end{figure*}

Figures \ref{fig:Alpha_TotalFailuresBetween5and30}
and \ref{fig:F_TotalFailuresBetween5and30}
aggregate the relative errors across all the datasets, for \alfa\, and \favg, respectively.
The proportion of errors is consistent between \alfa\, and \favg, but there are
more large errors when the performance is measured by \alfa. 
This is due to smaller error magnitude when the performance is measured by \alfa\, 
in contrast to \favg, since \alfa\, takes classification by chance into account.
With respect to individual estimation procedures, there is a considerable 
divergence of the random cross-validation.
For both performance measures, \alfa\, and \favg, it consistently incurs higher 
proportion of large errors and lower proportion of small errors in comparison to
the rest of the estimation procedures.

\FloatBarrier

\subsection{Friedman test} 
\label{sec:friedman}

The Friedman test is used to compare multiple procedures over multiple datasets
\cite{friedman37,friedman40,iman80,demsar2006statistical}.
For each dataset, it ranks the procedures by their performance.
It tests the null hypothesis that the average ranks of the procedures across all 
the datasets are equal. If the null hypothesis is rejected, one applies 
the Nemenyi post-hoc test \cite{nemenyi63} on pairs of procedures.
The performance of two procedures is significantly different if their average
ranks differ by at least the critical difference.
The critical difference depends on the number of procedures to compare,
the number of different datasets, and the selected significance level.

\begin{figure*}[h!]
\begin{center}
\includegraphics[width=\textwidth]{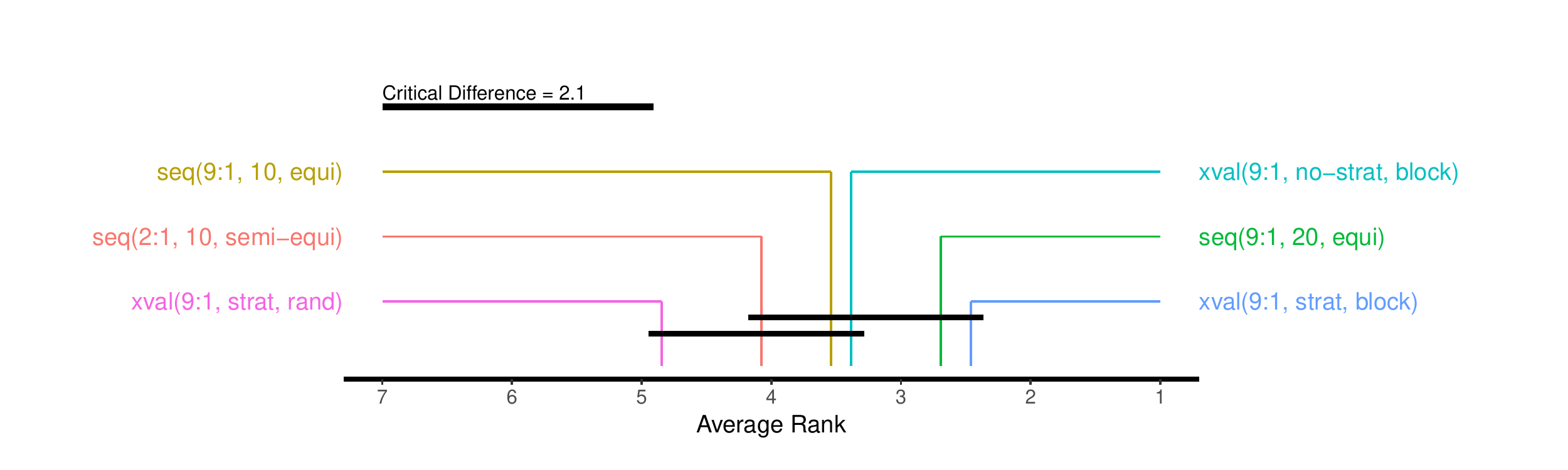}
\caption{Ranking of the six estimation procedures according to the Friedman-Nemenyi test.
The average ranks are computed from absolute errors, measured by \alfa.
The black bars connect ranks that are not significantly different at the 5\% level.}
\label{fig:FriedmanAlpha}
\end{center}
\end{figure*}

\begin{figure*}[h!]
\begin{center}
\includegraphics[width=\textwidth]{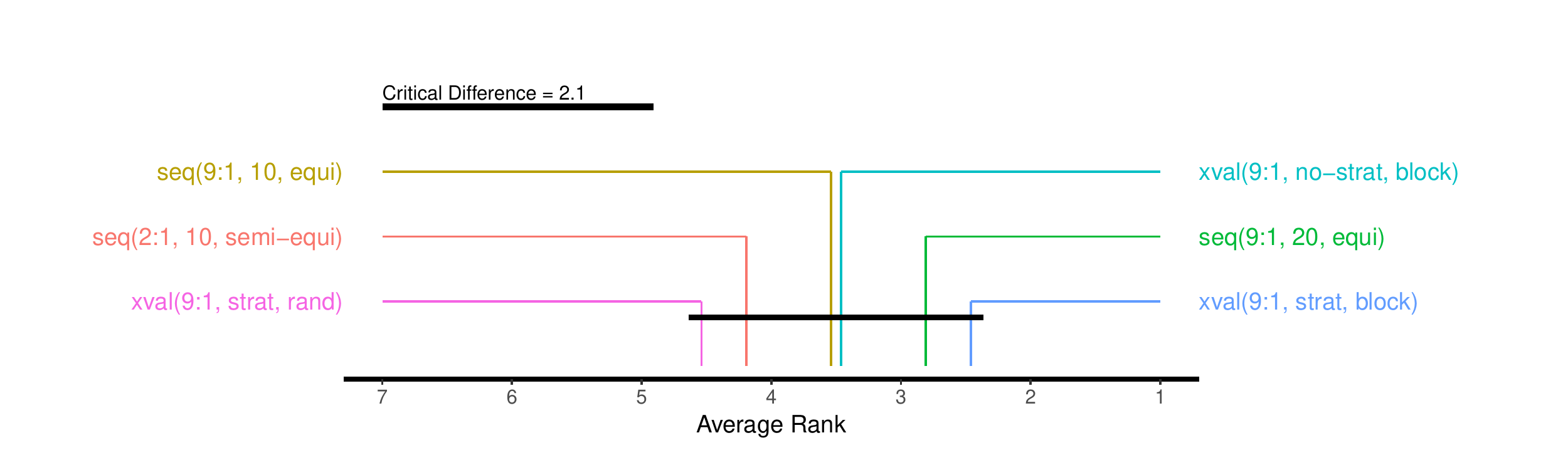}
\caption{Ranking of the six estimation procedures according to the Friedman-Nemenyi test.
The average ranks are computed from absolute errors, measured by \favg.
The black bar connects ranks that are not significantly different at the 5\% level.}
\label{fig:FriedmanF1}
\end{center}
\end{figure*}

In our case, the performance of an estimation procedure is taken as the absolute
error it incurs: $AbsErr = |Est - Gold|$.
The estimation procedure with the lowest absolute error gets the lowest (best) rank.
The results of the Friedman-Nemenyi test are in
Figures \ref{fig:FriedmanAlpha} 
and \ref{fig:FriedmanF1}, for \alfa\, and \favg, respectively.

For both performance measures, \alfa\, and \favg, the Friedman rankings are the same.
For six estimation procedures, 13 language datasets, and the 5\% significance level,
the critical difference is $2.09$.
In the case of \favg\, (Figure \ref{fig:FriedmanF1})
all six estimation procedures are within the critical difference,
so their ranks are not significantly different.
In the case of \alfa\, (Figure \ref{fig:FriedmanAlpha}), however,
the two best methods are significantly better than the random cross-validation.

\FloatBarrier

\subsection{Wilcoxon test}
\label{sec:wilcoxon}

The Wilcoxon signed-rank test is used to compare two procedures on related data
\cite{wilcoxon1945,demsar2006statistical}.
It ranks the differences in performance of the two procedures, 
and compares the ranks for the positive and negative differences.
Greater differences count more, but the absolute magnitudes are ignored.
It tests the null hypothesis that the differences follow a symmetric distribution around zero.
If the null hypothesis is rejected one can conclude that one procedure outperforms
the other at a selected significance level.

\begin{figure*}[h!]
\begin{center}
\includegraphics[width=\textwidth]{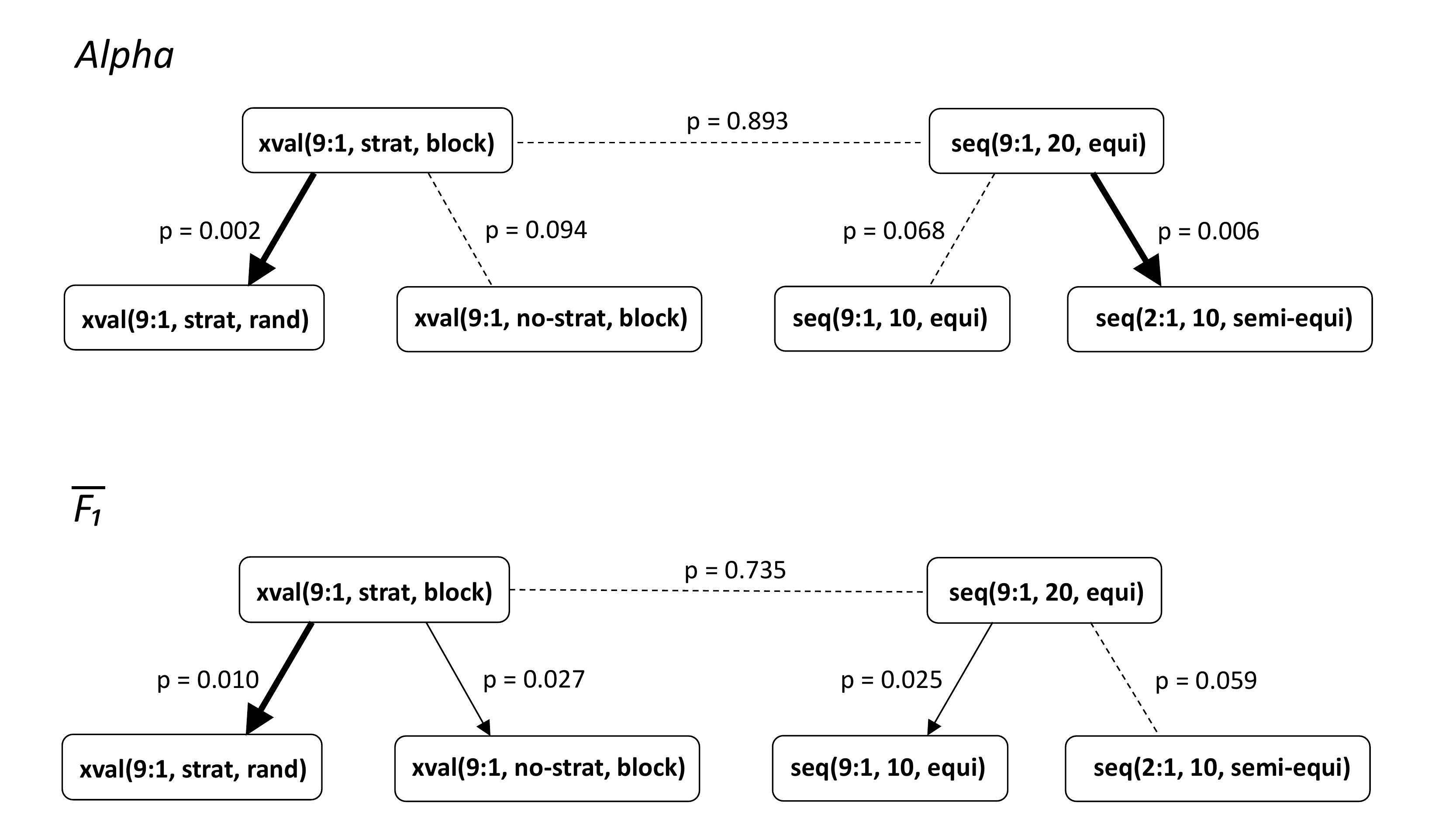}
\caption{Differences between pairs of estimation procedures according to the Wilcoxon
signed-rank test. Compared are the average absolute errors, measured by \alfa\, (top) 
and \favg\, (bottom). Thick solid lines denote significant differences at the 
1\% level, normal solid lines significant differences at the 5\% level, 
and dashed lines insignificant differences.
Arrows point from a procedure which incurs smaller errors to a procedure
with larger errors.}
\label{fig:Wilcoxon}
\end{center}
\end{figure*}

In our case, the performance of pairs of estimation procedures is compared at the
level of language datasets.
The absolute errors of an estimation procedure are averaged across the in-sets 
of a language. The average absolute error is then $AvgAbsErr = \sum |Est - Gold| / L$, 
where $L$ is the number of in-sets.
The results of the Wilcoxon test, for selected pairs of estimation procedures,
for both \alfa\, and \favg, are in Figure \ref{fig:Wilcoxon}.

The Wilcoxon test results confirm and reinforce the main results of the previous
sections. Among the cross-validation procedures, blocked cross-validation is
consistently better than the random cross-validation, at the 1\% significance level. 
Stratified approach is better than non-stratified, but significantly 
(5\% level) only for \favg. 
The comparison of the sequential validation procedures is less conclusive.
The training:test set ratio 9:1 is better than 2:1, but significantly (at the 5\% level)
only for \alfa. With the ratio 9:1 fixed, 20 samples yield better performance
estimates than 10 samples, but significantly (5\% level) only for \favg.
We found no significant difference between the best cross-validation and sequential 
validation procedures in terms how well they estimate the average absolute errors.

\FloatBarrier

\section{Conclusions}
\label{sec-conclusions}

In this paper we present an extensive empirical study about the performance 
estimation procedures for sentiment analysis of Twitter data.
Currently, there is no settled approach on how to properly evaluate models in
such a scenario. Twitter time-ordered data shares some properties of
static data for text mining, and some of time series data. 
Therefore, we compare estimation procedures developed for both types of data.

The main result of the study is that standard, random cross-validation
should not be used when dealing with time-ordered data. Instead, one should
use blocked cross-validation, a conclusion already corroborated
by Bergmeir et al. \cite{bergmeir2011forecaster,Bergmeir2012}. 
Another result is that we find no significant differences
between the blocked cross-validation and the best sequential validation.
However, we do find that cross-validations typically overestimate
the performance, while sequential validations underestimate it.

The results are robust in the sense that we use two different performance measures,
several comparisons and tests, and a very large collection of data.
To the best of our knowledge, we analyze and provide by far the largest 
set of manually sentiment-labeled tweets publicly available.

There are some biased decisions in our creation of the gold standard though,
which limit the generality of the results reported, and
should be addressed in the future work.
An out-set always consists of 10,000 tweets, and immediately follows the in-sets.
We do not consider how the performance drops over longer out-sets,
nor how frequently should a model be updated. More importantly, we
intentionally ignore the issue of dependent observations,
between the in- and out-sets, and between the training and test sets.
In the case of tweets, short-term dependencies are demonstrated in the form 
of retweets and replies.
Medium- and long-term dependencies are shaped by periodic events, influential users and communities, 
or individual user's habits. When this is ignored, the model performance is
likely overestimated. Since we do this consistently, our comparative results still hold.
The issue of dependent observations was already addressed for blocked cross-validation 
\cite{racine2000consistent,bergmeir2014} by removing adjacent observations
between the training and test sets, thus effectively creating a gap between the two. 
Finally, it should be noted that different Twitter language datasets are of different 
sizes and annotation quality, belong to different time periods, and that there are time periods 
in the datasets without any manually labeled tweets.

\section*{Data and code availability}

All Twitter data were collected through the public Twitter API and
are subject to the Twitter terms and conditions.
The Twitter language datasets are available in a public language resource repository 
\textsc{clarin.si} at \url{http://hdl.handle.net/11356/1054}, and are described 
in \cite{mozetic2016multilingual}.
There are 15 language files, where the Serbian/Croatian/Bosnian dataset is
provided as three separate files for the constituent languages.
For each language and each labeled tweet, there is the tweet ID
(as provided by Twitter), the sentiment label
(negative, neutral, or positive), and the annotator ID (anonymized). 
Note that Twitter terms do not allow to openly publish the original tweets, 
they have to be fetched through the Twitter API.
Precise details how to fetch the tweets, given tweet IDs, are provided
in Twitter API documentation
\url{https://developer.twitter.com/en/docs/tweets/post-and-engage/api-reference/get-statuses-lookup}.
However, upon request to the corresponding author, a bilateral agreement on the
joint use of the original data can be reached.

The TwoPlaneSVMbin classifier and several other machine learning algorithms
are implemented in an open source LATINO library \cite{Grcar2015phd}.
LATINO is a light-weight set of software components for building text mining 
applications, openly available at \url{https://github.com/latinolib}.

All the performance results, for gold standard and the six estimation procedures, 
are provided in a form which allows for easy reproduction
of the presented results. The \textbf{R} code and data files needed to reproduce all 
the figures and tables in the paper are available at 
\url{http://ltorgo.github.io/TwitterDS/}.

\section*{Acknowledgements}

Igor Mozeti\v{c} and Jasmina Smailovi\'{c} acknowledge financial support from the 
H2020 FET project DOLFINS (grant no. 640772),
and the Slovenian Research Agency (research core funding no. P2-0103). 

Luis Torgo and Vitor Cerqueira acknowledge financing by project 
``Coral - Sustainable Ocean Exploitation: Tools and
Sensors/NORTE-01-0145-FEDER-000036'', financed by the North Portugal Regional
Operational Programme (NORTE 2020), under the PORTUGAL 2020 Partnership
Agreement, and through the European Regional Development Fund (ERDF).

We thank Miha Gr\v{c}ar and Sa\v{s}o Rutar for valuable discussions and
implementation of the LATINO library.


%
%
%

\end{document}